\documentclass[letterpaper]{article}

\usepackage[utf8]{inputenc}
\usepackage{mathtools}
\usepackage{listings}
\usepackage[algo2e,linesnumbered,ruled,lined]{algorithm2e}
\usepackage{amsmath}
\usepackage{amssymb} 
\usepackage{dsfont}
\usepackage{graphicx}
\usepackage{subfig}
\usepackage{aaai}
\usepackage{times}
\usepackage{helvet}
\usepackage{courier}
\usepackage{xcolor}
\usepackage{multirow}
\usepackage{multicol}
\usepackage{hyperref}
\hypersetup{
	colorlinks=true,
	linkcolor=black,
	filecolor=black,      
	urlcolor=black,
	citecolor=black,}
\usepackage{tikz}
\tikzset{>=latex}
\usetikzlibrary{automata,positioning}

\newtheorem{definition}{Definition}
\newtheorem{remark}{Remark}
\begin{document}

\title{Modular Deep Reinforcement Learning with Temporal Logic Specifications}
\author{Lim Zun Yuan, Mohammadhosein Hasanbeig, Alessandro Abate, and Daniel Kroening \\Department of Computer Science, University of Oxford \\Parks Road, Oxford, OX1 3QD\\}
\maketitle
\begin{abstract}
\begin{quote}
We propose an actor-critic, model-free, and online Reinforcement Learning (RL) framework for continuous-state continuous-action Markov Decision Processes (MDPs) when the reward is highly sparse but encompasses a high-level temporal structure. We represent this temporal structure by a finite-state machine and construct an on-the-fly synchronised product with the MDP and the finite machine. The temporal structure acts as a guide for the RL agent within the product, where a modular Deep Deterministic Policy Gradient (DDPG) architecture is proposed to generate a low-level control policy. We evaluate our framework in a Mars rover experiment and we present the success rate of the synthesised policy. 
\end{quote}
\end{abstract}

\maketitle

\section{Introduction}

{Deep reinforcement learning} is an emerging paradigm for autonomous solving of decision-making tasks in complex and unknown environments. However, tasks featuring extremely delayed rewards are often difficult, if at all possible, to solve with monolithic learning in Reinforcement Learning (RL). A well-known example is the Atari game Montezuma's Revenge in which deep RL methods such as \cite{deepql} failed to score even once. 

Despite their generality, it is not fair to compare deep RL methods with how humans learn these problems, since humans already have prior knowledge and associations regarding elements and their corresponding function, e.g. ``keys open doors'' in Montezuma's Revenge. These simple yet critical temporal high-level associations in Montezuma's Revenge and a large number of real world complex problems, can lift deep RL initial knowledge about the problem to efficiently find the global optimal policy, while avoiding an exhaustive unnecessary exploration in the beginning. 

These hierarchies, sometimes called options \cite{sutton}, can be encoded in general RL algorithms to solve such complex problems. Practical approaches in hierarchical RL depend on state representations and on whether they are simple or structured enough such that suitable reward signals can be effectively engineered by hand. This means that these methods often require detailed supervision in the form of explicitly specified high-level actions or intermediate supervisory signals \cite{precup,options_1,options_h_1,options_2,options_h_2,modular}. 

In this paper we propose a fully-unsupervised one-shot online learning framework for deep RL, where the learner is presented with a composable {high-level} mission task in a continuous-state and continuous-action MDP. The mission task is specified in the form of Linear Temporal Logic (LTL) property, namely a formal, un-grounded, and symbolic representation of the task and of its components. {Without requiring any supervision}, each component of the LTL property {systematically} structures any complex mission task into low-level, achievable task ``modules". The given LTL property essentially acts as a high-level and unsupervised guide for the agent, whereas the low-level planning is done by a deep RL scheme.

{LTL is a rich specification language that} can formally express a wide range of time-dependent logical properties which are quite similar to patterns in natural language \cite{natural2LTL,natural2LTL2,natural2LTL3}. Examples include safety, liveness and cyclic properties, where the agent is required to make progress (liveness) while executing components for critical sections (safety) or to perform a sequence of tasks periodically (cyclic). 

In order to synchronise the high-level LTL guide with RL, we convert the LTL property to an automaton, namely a finite-state machine accepting sequences of symbols \cite{bible}. Once the automaton is generated from the LTL property, we construct on-the-fly a synchronous product between the MDP and the automaton\footnote{On-the-fly here means that the agent can track the state of the underlying MDP and of the automaton without explicitly constructing the synchronous product.} and then automatically define a reward function based on the structure of the automaton. From this algorithmic reward-shaping procedure, an RL agent is able to accomplish highly complex tasks with no supervisory assistance. 

The closest line of work is the model-based \cite{topku,dorsa} or model-free \cite{logicalconstraint,bolt,toro} approaches in RL that constrain the agent with a temporal logic property. {However, these approaches are limited to finite-state finite-action MDPs, an assumption we will relax throughout this work.} Another related work is \cite{modular}'s policy-sketch-based method, which learns easy instruction-based tasks first and  eventually composes them together, to accomplish a more complex task. {In this work instead, the  complex task can be expressed as an LTL property to guide the learning and to generate a policy with no need to start from easy tasks and later join them together}.

In addition, conventional RL is mostly focused on problems in which the set of states of the Markov Decision Processes (MDP) and the set of possible actions are finite. In \cite{wolf}, the property of interest is expressed via LTL, which is converted to a Deterministic Rabin Automaton (DRA). A modified Dynamic Programming (DP) method is applied, maximising the worst-case probability of satisfying the property. However, \cite{wolf} assumes to know the MDP a priori. \cite{topku,brazdil,dorsa} further assume that the given MDP has unknown transition probabilities and builds a Probably Approximately Correct MDP (PAC MDP), which is multiplied by the LTL property after conversion to DRA. The PAC MDP is generated via an RL-like algorithm. Nevertheless, many real world problems require {continuous real-valued} actions to be taken in response to high-dimensional and real-valued state observations. 

Unfortunately, for MDPs with continuous state and continuous action spaces, to the best of our knowledge, no research has been done on the problem of policy synthesis under full LTL in RL. To tackle problems with continuous state and action spaces in RL, the most immediate method is to discretise the state- and action spaces of the MDP \cite{abate2010approximate,abate2015quantitative} and to rely on conventional deep RL methods. Although this discretisation method works well for many problems \cite{faust,stochy}, the produced discrete MDP might be approximate and might not capture the full dynamics of the original MDP, which can be essential for optimally solving the original problem. Further, the number of discrete actions increases exponentially with the number of degrees of freedom \cite{DDPG} - a similar consideration holds for the state space. Thus, discretisation of MDPs generally suffers from the trade off between accuracy and the curse of dimensionality. 

To tackle this issue, in this work we propose a modular Deep Deterministic Policy Gradient (DDPG) based on the results in \cite{DPG,DDPG}. This modular DDPG is the first actor-critic algorithm using deep function approximators that can learn policies in continuous action and state spaces while jointly optimises over LTL task-specific sub-policies. In summary, the contributions of this work can be listed as follows:
\begin{itemize}
    \item Our approach can deal with continuous-state, continuous-action, and fully-unknown MDPs. The proposed algorithm is the first LTL-guided model-free RL that can handle such MDPs and as a result, significantly increases the scalability and applicability of LTL-synthesis solutions.
    \item Curriculum learning approaches such as \cite{modular} need to learn easy sub-tasks first and then stitch the trained deep net policies together. Whereas, the proposed method learns and stitches sub-tasks together simultaneously in a one-shot learning scenario.
    \item Full LTL is an infinite-time horizon language, with which we can express a wide range of important properties that cannot be specified otherwise. Such properties include but are not limited to surveillance (e.g. repeatedly visiting certain locations in a given order while avoiding certain locations) or global objectives (e.g. always keeping energy level above threshold).
    \item The proposed reward function in this work is automatically shaped on-the-fly with no supervision through, allowing us to also automatically modularise global complex task into easy sub-tasks.
\end{itemize}

The rest of this article is organised as follow: First we reviews basic concepts and definitions. We then formally discuss the policy synthesis problem that we are dealing with, and we propose a modular deep RL method to constrain it. Lastly, case studies are provided to quantify the performance of the proposed algorithm.

\section{Problem Framework}\label{background}
\begin{definition}[General MDP]\label{mdpdef} The tuple $\mathfrak{M}=(\mathcal{S}, \allowbreak \mathcal{A}, \allowbreak s_0,\allowbreak P,\allowbreak \mathcal{AP},\allowbreak L)$ is a general MDP over a set of continuous states $\mathcal{S}=\mathds{R}^n$, where $\mathcal{A}=\mathds{R}^m$ is a set of continuous actions, and $s_0 \in \mathcal{S}$ is the initial state. $P:\mathcal{B}(\mathds{R}^n)\times\mathcal{S}\times\mathcal{A}\rightarrow [0,1]$ is a Borel-measurable conditional transition kernel which assigns to any pair of state $s \in \mathcal{S}$ and action $a \in \mathcal{A}$ a probability measure $P(\cdot|s,a)$ on the Borel space $(\mathds{R}^n,\mathcal{B}(\mathds{R}^n))$. $\mathcal{AP}$ is a finite set of atomic propositions and a labelling function $L: \mathcal{S} \rightarrow 2^{\mathcal{AP}}$ assigns to each state $s \in \mathcal{S}$ a set of atomic propositions $L(s) \subseteq 2^\mathcal{AP}$ \cite{shreve}.
\end{definition}

\begin{definition}[Path] 
An infinite path $\rho$ starting at $s_0$ is a sequence of states $\rho= s_0 \xrightarrow{a_0} s_1 \xrightarrow{a_1} ... ~$ such that every transition $s_i \xrightarrow{a_i} s_{i+1}$ is allowed in $\mathfrak{M}$, i.e. $s_{i+1}$ belongs to the smallest Borel set $B$ such that $P(B|s_i,a_i)=1$. 
\end{definition}

At each state $s \in \mathcal{S}$, an agent behaviour is determined by a Markov policy $\pi$, which is a mapping from states to a probability distribution over
the actions, i.e. $\pi: \mathcal{S} \rightarrow \mathcal{P}(\mathcal{A})$. {If $\mathcal{P}(\mathcal{A})$ is a degenerate distribution then the policy $\pi$ is said to be  deterministic.}

\begin{definition} [Expected Discounted Reward] 
	\label{expectedut}
	For a policy $\pi$ on an MDP $\mathfrak{M}$, the expected discounted reward is defined as \cite{sutton}:
	\begin{equation}
	\label{expecteduteq}
	{U}^{\pi}(s)=\mathds{E}^{\pi} [\sum\limits_{n=0}^{\infty} \gamma^n~ R(s_n,a_n)|s_0=s],
	\end{equation}
	where $\mathds{E}^{\pi} [\cdot]$ denotes the expected value given that the agent follows policy $\pi$, $\gamma\in [0,1]$ is a discount factor, $R:\mathcal{S}\times\mathcal{A}\rightarrow \mathds{R}$ is the reward, and $s_0,a_0,...,s_n,a_n$ is the sequence of state-action pairs generated by policy $\pi$ up to time step $n$. 
\end{definition}

{The function ${U}^{\pi}(s)$ is often referred to as value function (under the policy $\pi$). Another closely related notion in RL is action-value function $Q^\pi(s,a)$, which describes the expected discounted reward after taking an action $a$ in state $s$ and thereafter following policy $\pi$:
$$
Q^\pi(s,a)=\mathds{E}^{\pi} [\sum\limits_{n=1}^{\infty} \gamma^n~ R(s_n,a_n)|s_0=s,a_0=a].
$$
Accordingly, the recursive form of the action-value function can be obtained as:}
\begin{equation}\label{eq_qrecursive}
    Q^\pi(s,a)=R(s,a)+\gamma Q^\pi(s_1,a_1), 
\end{equation}
where $a_1 = \pi(s_1)$. 
Q-learning (QL) \cite{watkins} is the most extensively used model-free RL algorithm {built upon \eqref{eq_qrecursive}}, for MDPs with finite-state and finite-action spaces. For all state-action pairs QL initializes a Q-function $Q^\beta(s,a)$ with an arbitrary finite value, {where $\beta$ is an arbitrary  stochastic policy}. 
{QL is an off-policy RL scheme, namely policy $\beta$ has no effect on the convergence of the Q-function}, as long as every state-action pair is visited infinitely many times. {Thus, for the sake of simplicity, we may drop the policy index $\beta$ from the action-value function}. Under mild assumptions, QL converges to a unique limit, and a greedy policy $\pi^*$ can be obtained as follows:
$$
\pi^*(s)=\arg\max\limits_{a \in \mathcal{A}}~Q(s,a),
$$
and $ \pi^* $ corresponds to the optimal policy that is generated DP \cite{NDP} to maximise \eqref{expecteduteq}, when the MDP is fully known. 

The DPG algorithm \cite{DPG} introduces a parameterised function $\mu(s|\theta^\mu)$ called actor to represent the current policy by deterministically mapping states to actions, where $\theta^\mu$ is the function approximation parameters for the actor function. Further, an action-value function $Q(s,a|\theta^Q)$ is called critic and is learned as described next.  

{Assume that at time step $t$ the agent is at state $s_t$, takes action $a_t$, and receives a scalar reward $R(s_t,a_t)$. In case when the agent policy is deterministic,} the recursion \eqref{eq_qrecursive} can be approximated by parameterising $Q$ using {a parameter set} $\theta^Q$, {i.e. $Q(s_t, a_t|\theta^Q)$}, and by minimizing the following loss function:
\begin{equation}\label{critic}
    L(\theta^Q)= \mathbb{E}^\pi_{s_t \sim \rho^\beta}[(Q(s_t, a_t|\theta^Q)-y_t)^2],
\end{equation}
where $\rho^\beta$ is {the probability distribution of state visits over $\mathcal{S}$, under} any given arbitrary stochastic policy $\beta$, and $y_t = R(s_t,a_t)+\gamma Q(s_{t+1}, a_{t+1}|\theta^Q)$ such that $a_{t+1}=\pi(s_{t+1})$.

The actor is updated by applying the chain rule to the expected return with respect to the actor parameters as follows: 
\begin{equation}
    \begin{aligned}
    &\nabla_{\theta^\mu} U^\mu(s_t) \approx \mathbb{E}_{s_t\sim p^\beta}[\nabla_{\theta^\mu} Q(s,a|\theta^Q)|_{s=s_t, a=\mu(s_t|\theta^\mu)}]  \\
    &= \mathbb{E}_{s_t\sim p^\beta}[\nabla_{a} Q(s,a|\theta^Q)|_{s=s_t, a=\mu(s_t)}\nabla_{\theta^\mu}\mu(s|\theta^\mu)|_{s=s_t}].
    \end{aligned}
\end{equation}
\cite{DPG} has shown that this is a policy gradient, and therefore we can apply a policy gradient algorithm on the deterministic policy. {DDPG further extends DPG by employing a deep neural network as function approximator and updating the network parameters via a ``soft update'' method, which is explained later in the paper.} 

\subsection{Linear Temporal Logic (LTL)}
We employ LTL to encode the structure of the high-level mission task and to automatically shape the reward function. An LTL formula is able to express a range of properties that are hard (if at all possible) to express by conventional or handcrafted methods in classical reward shaping \cite{sutton,precup,options_h_2}. LTL formulae $\varphi$ over a given set of atomic propositions $\mathcal{AP}$ are syntactically defined as \cite{pnueli}
\begin{equation}\label{ltlsyntax}
	\varphi::= true ~|~ \alpha \in \mathcal{AP} ~|~ \varphi \land \varphi ~|~ \neg \varphi ~|~ \bigcirc \varphi ~|~ \varphi \cup \varphi,
\end{equation}
where the operators $ \bigcirc $ and $ \cup $ are called ``next'' and ``until'', respectively.

We will next define the semantics of LTL formulae interpreted over MDPs. For a given path $\rho$, we define the $i$-th state of $\rho$ to be $\rho[i]$ where $\rho[i] = s_{i}$, and the $i$-th suffix of $\rho$ to be $\rho[i..]$ where $\rho[i..]=s_i\xrightarrow{a_i} s_{i+1} \xrightarrow{a_{i+1}} s_{i+2} \ldots~$

\begin{definition}
	[LTL Semantics] \label{semantics} 
	For an LTL formula $\varphi$ and for a path $\rho$, the satisfaction relation $\rho\models\varphi$ is defined as
	\begin{equation*}
	\resizebox{0.98\columnwidth}{!}{$
	\begin{aligned}
	& \rho \models \alpha \in \mathcal{AP} \Leftrightarrow \alpha \in L(\rho[0]), \\
	& \rho \models \varphi_1\wedge \varphi_2 \Leftrightarrow \rho \models \varphi_1\wedge \rho \models \varphi_2,\\
	& \rho \models \neg \varphi \Leftrightarrow \rho \not \models \varphi, \\
	& \rho \models \bigcirc \varphi \Leftrightarrow \rho[1..] \models \varphi, \\
	& \rho \models \varphi_1\cup \varphi_2 \Leftrightarrow \exists j \geq 0 : \rho[j..] \models \varphi_2 \wedge\forall i, 0 \leq i < j, \rho[i..] \models \varphi_1.
	\end{aligned}$}
	\end{equation*} 
\end{definition}

The operator next \resizebox{3.3mm}{!}{$\bigcirc$} requires that $\varphi$ to be satisfied starting from the next-state suffix of $\rho$. The operator until $\cup$ is satisfied over $\rho$ if $\varphi_1$ continuously holds until $\varphi_2$ becomes true. Using the until operator $\cup$ we can define two temporal modalities: (1)
eventually, $\lozenge \varphi = true \cup \varphi$; and (2) always, $\square
\varphi = \neg \lozenge \neg \varphi$. LTL extends propositional logic using the temporal modalities until $ \cup $, eventually $ \lozenge $, and always $ \square $. For instance, constraints such as ``eventually reach this point", ``visit these points in a particular sequential order", or ``always stay safe" are easily expressible by these modalities. Further, these modalities can be combined with logical connectives and nesting to provide more complex task specifications. Any LTL task specification $\varphi$ over $\mathcal{AP}$ expresses the following set of words:
$$
\mathit{Words}(\varphi)=\{\sigma \in (2^{\mathcal{AP}})^\omega ~\mbox{s.t.}~ \sigma \models \varphi\}.
$$
\begin{definition}[LTL Policy Satisfaction]
	We say that a stationary deterministic policy $ \pi $ satisfies an LTL formula $ \varphi $ if
	$
	\mathds{P}[L(s_0)L(s_1)L(s_2)...\allowbreak\in \mathit{Words}(\varphi)] \neq 0,
	$
	where every transition $s_i \rightarrow s_{i+1},~i=0,1,...$ is executed by taking action $ \pi(s_i) $ at state $ s_i $.  
\end{definition}

The set of associated words $\mathit{Words}(\varphi)$ is expressible using a finite-state machine \cite{bible}. Limit Deterministic B\"uchi Automaton (LDBA) \cite{sickert} is the state-of-the-art in formal methods and proved to be the most succinct finite-state machine for this purpose \cite{sickert2}. We first define a Generalized B\"uchi Automaton (GBA), then we formally introduce
the LDBA.

\begin{definition}[Generalized B\"uchi Automaton] \label{gba_definition}
	A~GBA $\mathfrak{A}\allowbreak=(\allowbreak\mathcal{Q},\allowbreak q_0,\allowbreak\Sigma, \allowbreak\mathcal{F}, \allowbreak\Delta)$ is a state machine,  where $\mathcal{Q}$ is a finite set of states, $q_0 \subseteq \mathcal{Q}$ is the set of initial states, $\Sigma=2^\mathcal{AP}$ is a finite alphabet, $\mathcal{F}=\{F_1,...,F_f\}$ is the set of accepting conditions where $F_j \subset \mathcal{Q}, 1\leq j\leq f$, and $\Delta: \mathcal{Q} \times \Sigma \rightarrow 2^\mathcal{Q}$ is a transition relation.  
\end{definition}

Let $\Sigma^\omega$ be the set of all
infinite words over $\Sigma$. An infinite word $w \in \Sigma^\omega$ is
accepted by a GBA $\mathfrak{A}$ if there exists an infinite run $\theta \in
\mathcal{Q}^\omega$ starting from $q_0$ where $\theta[i+1] \in
\Delta(\theta[i],\omega[i]),~i \geq 0$ and, for each $F_j \in \mathcal{F}$,

\begin{equation} \label{acc}
\mathit{inf}(\theta) \cap F_j \neq \emptyset, 
\end{equation}

where $\mathit{inf}(\theta)$ is the set of states that are visited
infinitely often in the sequence $\theta$. 

\begin{definition}
	[LDBA]
	\label{ldbadef}
	A GBA $\mathfrak{A}=(\mathcal{Q},q_0,\Sigma, \mathcal{F}, \Delta)$ is
	limit deterministic if $\mathcal{Q}$ can be partitioned into two disjoint sets $\mathcal{Q}=\mathcal{Q}_N \cup \mathcal{Q}_D$, 
	such that \cite{sickert}:
	\begin{itemize}
		\item $\Delta(q,\alpha) \subseteq \mathcal{Q}_D$ and $|\Delta(q,\alpha)|=1$ for every state $q\in\mathcal{Q}_D$ and for every corresponding $\alpha \in \Sigma$,
		\item for every $F_j \in \mathcal{F}$, $F_j \subset \mathcal{Q}_D$.
	\end{itemize}
\end{definition}
In other words, a LDBA is a GBA with two partitions: (1) initial ($\mathcal{Q}_N$), and (2) accepting ($\mathcal{Q}_D$). The accepting partition includes all accepting states and also all the transitions are deterministic. 

\begin{remark}
The standard method in almost all of the works on RL LTL policy synthesis in finite-state MDPs, is to convert the given LTL formula into a DRA, which is known that results in automata that are doubly exponential in the size of the given LTL formula. Conversely, LDBA, as the current state-of-the-art, is only an exponential-sized automaton for LTL$\setminus$GU (a fragment of LTL) and is the same size as deterministic automata for the rest of LTL \cite{sickert}. This dramatically decreases the size of the automaton for the same LTL formula \cite{sickert2}, and as we see later, significantly enhances the convergence rate of RL due to the smaller product state space (Definition \ref{product_mdp_def}). However, the use of LDBA in RL introduces non-trivial problems, such as partial non-determinism, to the learning process, which are addressed in the work. Furthermore, an LDBA is semantically easier than a DRA in terms of its acceptance conditions, which makes policy synthesis algorithms much simpler to implement.
\end{remark}

\begin{definition}[Non-accepting Sink Component]\label{nonacc}
	A non-accepting sink component of the LDBA $\mathfrak{A}$ is a directed graph induced by a set of states $ Q \subset\mathcal{Q}$ such that (1) the graph is strongly connected; (2) it does not include all accepting sets $ F_k,~k=1,...,f $; and (3) there exist no other strongly connected set $ Q' \subset \mathcal{Q},~Q'\neq Q $ such that $ Q \subset Q' $. We denote the union of all non-accepting sink components of $\mathfrak{A}$ as $\mathds{N}$. 
\end{definition}

The set $\mathds{N}$ include those components in the automaton that are surely non-accepting and impossible to escape from. Thus, reaching them is equivalent to not being able to satisfy the given LTL property.

\section{Modular Deep RL}\label{lcsec}

We consider a modular deep RL problem in which we exploit the structural information provided by the LTL specification and by constructing a sub-policy for each state of the associated LDBA. Our proposed approach learns a satisfying policy without requiring any information about the grounding of the LTL task to be explicitly specified. {Namely, the labelling {assignment} in Definition \ref{mdpdef} is unknown a-priori, and the algorithm solely relies on experience samples gathered on-the-fly.}  

Given an LTL mission task and an unknown continuous-state continuous-action MDP, we aim to synthesise a policy that satisfies the LTL specification. For the sake of clarity and to explain the core ideas of the algorithm, for now we assume that the MDP graph and the transition kernel are known: later these assumptions are entirely removed, and we stress that the algorithm can be run model-free. We relate the MDP and the automaton by synchronising them, in order to create a new structure that is first of all compatible with deep RL and secondly that encompasses the given logical property. 

\begin{definition} [Product MDP]\label{product_mdp_def}
	Given an MDP $\mathfrak{M}=(\allowbreak\mathcal{S},\allowbreak\mathcal{A},\allowbreak s_0,\allowbreak P,\allowbreak\mathcal{AP},L)$ and an LDBA $\mathfrak{A}=(\mathcal{Q},q_0,\Sigma, \mathcal{F}, \Delta)$ with $\Sigma=2^{\mathcal{AP}}$, the product MDP is defined as $(\mathfrak{M}\otimes \mathfrak{A}) = \mathfrak{M}_\mathfrak{A}=\mathcal{S}^\otimes,\allowbreak \mathcal{A},\allowbreak s^\otimes_0,P^\otimes,\allowbreak \mathcal{AP}^\otimes,\allowbreak L^\otimes,\allowbreak \mathcal{F}^\otimes)$, where $\mathcal{S}^\otimes = \mathcal{S}\times\mathcal{Q}$, $s^\otimes_0=(s_0,q_0)$, $\mathcal{AP}^\otimes = \mathcal{Q}$, $L^\otimes : \mathcal{S}^\otimes\rightarrow 2^\mathcal{Q}$ such that $L^\otimes(s,q)={q}$ and $\mathcal{F}^\otimes \subseteq {\mathcal{S}^\otimes}$ is the set of accepting states $\mathcal{F}^\otimes=\{F^\otimes_1,...,F^\otimes_f\}$, where ${F}^\otimes_j=\mathcal{S}\times F_j$. The transition kernel $P^\otimes$ is such that given the current state $(s_i,q_i)$ and action $a$, the new state is $(s_j,q_j)$, where $s_j\sim P(\cdot|s_i,a)$ and $q_j\in\Delta(q_i,L(s_j))$.
\end{definition}

By constructing the product MDP we synchronise the current state of the MDP with the state of the automaton. This allows to evaluate the (partial) satisfaction of the corresponding LTL property (or parts thereof), and consequently to {modularise} the high-level task into sub-tasks. Hence, with a proper reward assignment driven from the LTL property and its associated LDBA, the agent is able to break down a complex task into a set of easy sub-tasks. We elaborate further on task modularisation in the next section.

Note that the automaton transitions can be executed just by reading the label of the visited states, which makes the agent aware of the automaton state without explicitly constructing the product MDP. Thus, the proposed approach can run ``model-free'', and as such it does not require an initial knowledge about the MDP.

In the following we define an ``on-the-fly'' LTL-driven reward function, emphasising that the agent does not need to know the model structure or the transition probabilities (or their product). Before introducing a reward assignment for the RL agent, we need to present the ensuing function:
\begin{definition}
	[Accepting Frontier Function] \label{frontier} For~an~LDBA $\mathfrak{A}\allowbreak = (\allowbreak\mathcal{Q}, \allowbreak q_0, \allowbreak\Sigma, \allowbreak\mathcal{F}, \allowbreak\Delta)$, we define $ Acc:\mathcal{Q}\times 2^{\mathcal{Q}}\rightarrow2^\mathcal{Q} $ as the accepting frontier function, which executes the following operation over a given set $ \mathds{F}\subset 2^{\mathcal{Q}}$: 	
	\[Acc(q,\mathds{F})=\left\{
	\begin{array}{lr}
	\mathds{F}_{~\setminus F_j}~~~ & (q \in F_j) \wedge (\mathds{F}\neq F_j),\\
	\\
	{\{F_k\}_{k=1}^{f}}_{~\setminus F_j} ~~~ & (q \in F_j) \wedge (\mathds{F}=F_j),\\
	\\
	\mathds{F} & $otherwise.$
	\end{array}
	\right.
	\] 
\end{definition}

In words, once the state $ q\in F_j $ and the set $ \mathds{F} $ are introduced to the function $ Acc $, it outputs a set containing the elements of $ \mathds{F} $ minus $ F_j $. However, if $ \mathds{F}=F_j $, then the output is the family set of all accepting sets of the LDBA minus the set $ F_j $. Finally, if the state $ q $ is not an accepting state then the output of $ Acc $ is $ \mathds{F} $. The accepting frontier function excludes from $\mathds{F}$ the accepting set that is currently visited, unless it is the only remaining accepting set. Otherwise, the output of $Acc(q,\mathds{F})$ is $\mathds{F}$ itself. Owing to the automaton-driven structure of the $Acc$ function, we are able to shape a reward function (as detailed next) without any supervision and regardless of the dynamics of the MDP. 

We propose a reward function that observes the current state $ s^\otimes $, the current action $ a $, and the subsequent state $ {s^\otimes}' $, to provide the agent with a scalar value according to the current automaton state: 
\begin{equation}\label{thereward2}
\begin{aligned}
R(s^\otimes,a) = \left\{
\begin{array}{lr}
r_p &  if ~  q' \in \mathds{A},~{s^\otimes}'=(s',q'),\\
r_{n} & if ~ q' \in \mathds{N},~{s^\otimes}'=(s',q'),\\
0 &  otherwise.
\end{array}
\right.
\end{aligned}
\end{equation}  
Here $r_p$ is a positive reward and $r_n$ is a negative reward. A positive reward is assigned to the agent when it takes an action that leads to a state, the label of which is in $\mathds{A}$. The set $ \mathds{A} $ is called the accepting frontier set, is initialised as the family set $ \mathds{A}=\{F_k\}_{k=1}^{f} $, and is updated by the following rule every time after the reward function is evaluated: 
$$
\mathds{A}\leftarrow Acc(q',\mathds{A}). 
$$
The set $ \mathds{A} $ contains those accepting states that are visited at a given time. Thus, the agent is guided by the above reward assignment to visit these states and once all of the sets $ F_k,~k=1,...,f, $ are visited, the accepting frontier $ \mathds{A} $ is reset. As such, the agent is guided to visit the accepting sets infinitely often, and consequently, to satisfy the given LTL property. Finally, the set $\mathds{N}$ is the set of non-accepting sink components of the automaton, as per Definition \ref{nonacc}.

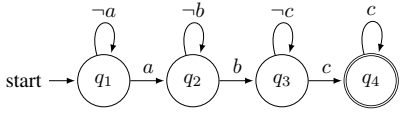
\begin{figure}[!t]\centering
	\scalebox{.79}
	    {
		\begin{tikzpicture}[shorten >=1pt,node distance=1.5cm,on grid,auto] 
		\node[state,initial] (q_1)   {$q_1$}; 
		\node[state] (q_2) [right=of q_1] {$q_2$};
		\node[state] (q_3) [right=of q_2] {$q_3$};
		\node[state,accepting] (q_4) [right=of q_3] {$q_4$};
		\path[->] 
		(q_1) edge [loop above] node {$\neg a$} ()
		(q_1) edge  node {$a$} (q_2)
		(q_2) edge [loop above] node {$\neg b$} ()
		(q_2) edge node {$b$} (q_3)
		(q_3) edge [loop above] node {$\neg c$} ()
		(q_3) edge node {$c$} (q_4)
		(q_4) edge [loop above] node {$c$} ();
		\end{tikzpicture}
		}
	\caption{LDBA for a sequential mission task.}
	\label{exam}
\end{figure}

\subsection{Task Modularisation}\label{segmentation}

In this section we explain how a complex task can be broken down into simple composable sub-tasks {or modules}. Each state of the automaton in the product MDP is a ``task divider'' and each transition between these states is a ``sub-task''. For example consider a sequential task of visit $a$ and then $b$ and finally $c$, i.e.
$$
\Diamond(a\wedge\Diamond (b \wedge\Diamond c)).
$$

The corresponding automaton for this LTL task is given in Fig. \ref{exam}. The entire task is modularised into three sub-tasks, i.e. reaching $a$, $b$, and then $c$, and each automaton state acts as a divider. 

{Given an LTL task and its LDBA $\mathfrak{A}\allowbreak = (\allowbreak\mathcal{Q}, \allowbreak q_0, \allowbreak\Sigma, \allowbreak\mathcal{F}, \allowbreak\Delta)$,} we propose a modular architecture of $n=|\mathcal{Q}|$ separate DDPG actor, actor-target, critic and critic-target neural networks, along with their own replay buffer. {A replay buffer is a finite-sized cache in which transitions sampled from exploring the environment are stored. The replay buffer is then used to train the actor and critic networks}. The set of neural nets acts as a global modular actor-critic deep RL architecture, which allows the agent to jump from one sub-task to another by just switching between the set of neural nets\footnote{Different embeddings, such as the one hot encoding \cite{onehot} and the integer encoding, have been applied in order to approximate the global Q-function with a single DDPG network. However, we have observed poor performance since these encodings allow the network to assume an ordinal relationship between automaton states. This means that by assigning integer numbers or one hot codes, automaton states are categorised in an ordered format, and can be ranked. Clearly, this disrupts Q-function generalisation by assuming that some states in the product MDP are closer to each other. Consequently, we have turned to the use of separate neural nets, which work together in a modular fashion, meaning that the agent can switch between these neural nets as it jumps from one automaton state to another.}. 

For each automaton state $q_i$ an actor function $\mu_{q_i}(s|\theta^{\mu_{q_i}})$ represents the current policy by deterministically mapping states to actions, where $\theta^{\mu_{q_i}}$ is the vector of parameters of the function approximation for the actor. The critic $Q_{q_i}(s,a|\theta^{Q_{q_i}})$ is learned based on \eqref{critic}, as in QL. 

The modular deep RL algorithm is detailed in Algorithm \ref{algor}. Each DDPG network set in this algorithm is associated with its own replay buffer $\mathcal{R}_{q_i}$, where $q_i \in \mathcal{Q}$ (line 4, 12). Experience samples are stored in $\mathcal{R}_{q_i}$ in the form of $$(s_i^\otimes, a_i, R_i, s_{i+1}^\otimes)=((s_i, q_i), a_i, R_i, (s_{i+1},\allowbreak q_{i+1}))$$. When the replay buffer reaches its maximum capacity, the samples are discarded based on a first in first out policy. At each time-step, actor and critic are updated by sampling a mini-batch of size $B$ uniformly from $\mathcal{R}_{q_i}$. We only train the DDPG network corresponding to the current automaton state, as experience samples on the current automaton state have little influence on other DDPG neural networks (line 12-17). 

{Further, directly implementing the update of the critic parameters as in \eqref{critic} is shown to be potentially unstable, and as a result the Q-update (line 14) is prone to divergence \cite{minhd}.} Hence, instead of directly copying the weights, the standard DDPG \cite{DDPG} uses ``soft'' target updates to improve learning stability. Target networks, $Q'$ and $\mu'$, are time-delayed copies of the original actor and critic networks that slowly track the learned networks, $Q$ and $\mu$. {These target actor and critic networks are used within the algorithm to gather evidence (line 13) and subsequently to update the actor and critic networks.} In our algorithm, for each automaton state $q_i$ we make a copy of the actor and the critic network: $\mu'_{q_i}(s|\theta^{\mu'_{q_i}})$ and $Q'_{q_i}(s,a|\theta^{Q'_{q_i}})$ respectively. The weights of both target networks are then updated by $\theta' = \tau \theta+(1-\tau)\theta'$ with a rate of $ \tau<1$ (line 18). {Although this ``soft update'' may slow down  learning as target networks have propagation delays, in practice this is greatly outweighed by the introduced learning stability.}

\begin{algorithm2e}[!t]
\DontPrintSemicolon
\SetKw{return}{return}
\SetKwRepeat{Do}{do}{while}
\SetKwData{conflict}{conflict}
\SetKwData{safe}{safe}
\SetKwData{sat}{sat}
\SetKwData{unsafe}{unsafe}
\SetKwData{unknown}{unknown}
\SetKwData{true}{true}
\SetKwInOut{Input}{input}
\SetKwInOut{Output}{output}
\SetKwFor{Loop}{Loop}{}{}
\SetKw{KwNot}{not}
\begin{small}
	\Input{LTL mission task $\varphi$}
	\Output{actor and critic networks}
	convert the LTL property $\varphi$ to an LDBA \;
	randomly initialise $|\mathcal{Q}|$ actors $\mu_i(s|\theta^{\mu_i})$ and critic $Q_i(s, a|\theta^{Q_i})$ networks with weights $\theta^{\mu_i}$ and $\theta^{Q_i}$, for each $q_i \in \mathcal{Q}$\;
	initialize $|\mathcal{Q}|$ corresponding target networks $\mu'_i$ and $Q'_i$ with weights $\theta^{\mu'_i} = \theta^{\mu_i}$, $\theta^{Q'_i}= \theta^{Q_i}$\;
	initialise $|\mathcal{Q}|$ replay buffers $\mathcal{R}_{i}$\;
	\Repeat{end of trial}
	{   
	    initialise $|\mathcal{Q}|$ random processes $\mathfrak{N}_i$\;
	    initialise state $s_1^\otimes=(s_1,q_1)$\;
		\For{$t=1$ \textbf{to} $max\_iteration\_number$}
		{   
		    choose action $a_t = \mu_{q_t}(s_t|\theta^{\mu_{q_t}})+\mathfrak{N}_{q_t}$ according to the current policy and exploration noise $\mathfrak{N}_i$\;
		    
		    execute action $a_t$ and observe reward 
    		$r_t$ and the new state $(s_{t+1}, q_{t+1})$\;
    		
    		store transition $((s_t,q_t), a_t, R_t, (s_{t+1}, q_{t+1}))$ in $\mathcal{R}_{q_t}$\;
    		
    		sample a random mini-batch of $B$ transitions 
    		$((s_i, q_i), a_i, R_i, (s_{i+1}, q_{i+1}))$ from $\mathcal{R}_{q_t}$\;
    		
    		set $y_i = R_i + \gamma Q_{q_{i+1}}'(s_{i+1}, \mu'_{q_{i+1}}(s_{i+1}|\theta^{\mu'_{q_{i+1}}})|\theta^{Q'_{q_{i+1}}})$\;
    		
    		update critic $Q_{q_t}$ and $\theta^{Q_{q_t}}$ by minimizing the loss:
    		$L =1/B\sum_{i}(y_i-Q_{q_t}(s_i, a_i|\theta^{Q_{q_t}}))^2$\;
    		
    		update the actor policy $\mu_{q_t}$ and $\theta^{\mu_{q_t}}$ by maximizing the sampled 
    		policy gradient:\;
    		$\nabla_{\theta^{\mu_{q_t}}} U^{\mu_{q_t}} \approx 1/B \sum_{i}[\nabla_{a} Q_{q_t}(s,a|\theta^{Q_{q_t}})|_{s=s_i, a=\mu_{q_t}(s_i|\theta^{\mu_{q_t}})}$\\ ~~$\nabla_{\theta^{\mu_{q_t}}}\mu_{q_t}(s|\theta^{\mu_{q_t}})|_{s=s_i}]$
    		
    		update the target networks:
    		$\theta^{Q'_{q_t}} \leftarrow \tau\theta^{Q_{q_t}} + (1-\tau)\theta^{Q'_{q_t}}$
    		$\theta^{\mu'_{q_t}} \leftarrow \tau\theta^{\mu^{q_t}} + (1-\tau)\theta^{\mu'_{q_t}}$\;
		}
	}
\end{small}
\caption{Modular Deep RL}
\label{algor}
\end{algorithm2e}

\section{Experiments}\label{case study}
In this section we discuss a mission planning problem for an autonomous Mars rover that uses the proposed algorithm to pursue exploration missions. The areas of interest on Mars are the Melas Chasma and the Victoria crater. 

\begin{figure}[!t]\centering
\includegraphics[width=0.7\columnwidth]{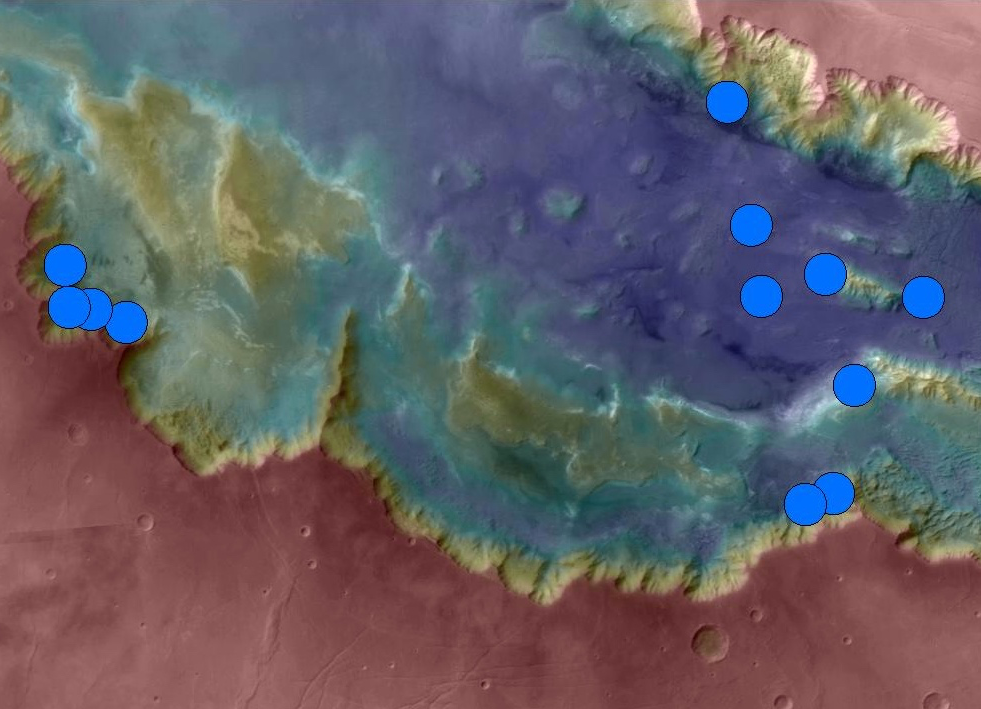}
\caption{Melas Chasma in the Coprates quadrangle, map color spectrum represents elevation, where red is high and blue is low. (Image courtesy of NASA, JPL, Caltech and University of Arizona)}
\label{Melas}
\end{figure}

The Melas Chasma a number of signs of water, with ancient river valleys and networks of stream channels showing up as sinuous and meandering ridges and lakes (Fig. \ref{Melas}). The blue dots, provided by NASA, indicate locations of Recurring Slope Lineae (RSL) in the canyon network. RSL are seasonal dark streaks regarded as the strongest evidence for the possibility of liquid water on the surface of Mars. RSL extend down-slope during a warm season and then disappear in the colder part of the Martian year \cite{water_on_mars}.

\begin{figure}[!t]
	\centering
	\subfloat[]{{\includegraphics[width=0.74\columnwidth]{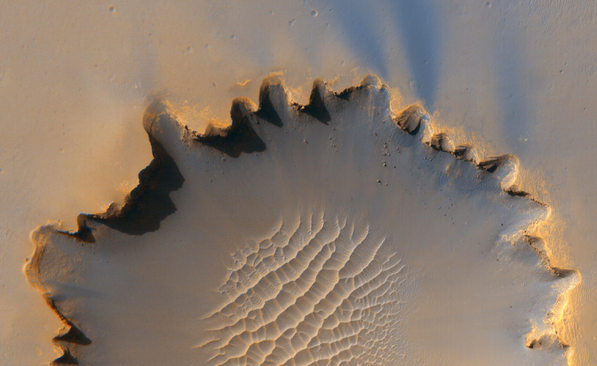} }}%
	\qquad
	\subfloat[]{{\includegraphics[width=0.74\columnwidth]{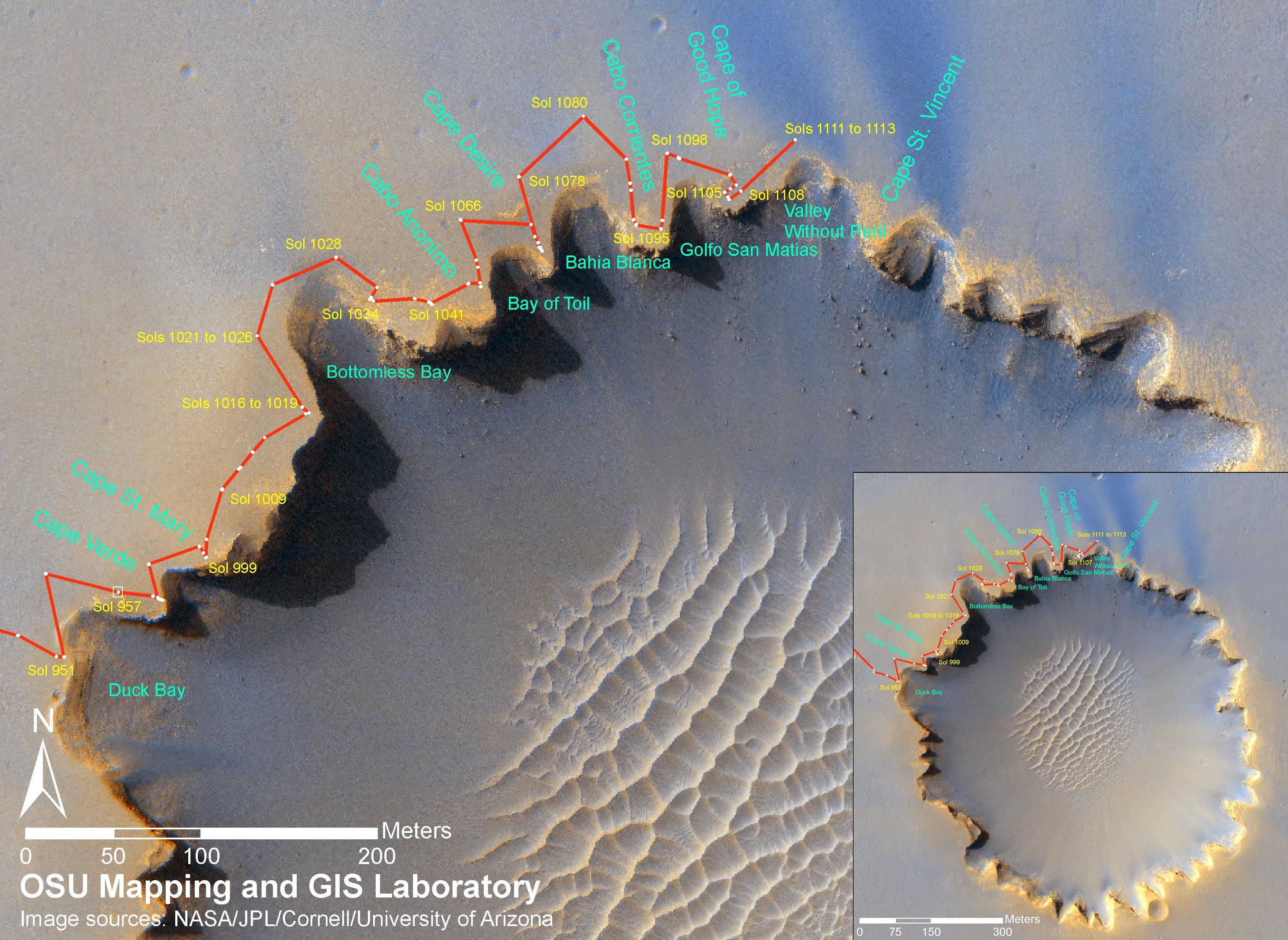} }}%
	\qquad
	\subfloat[]{{\includegraphics[width=0.74\columnwidth]{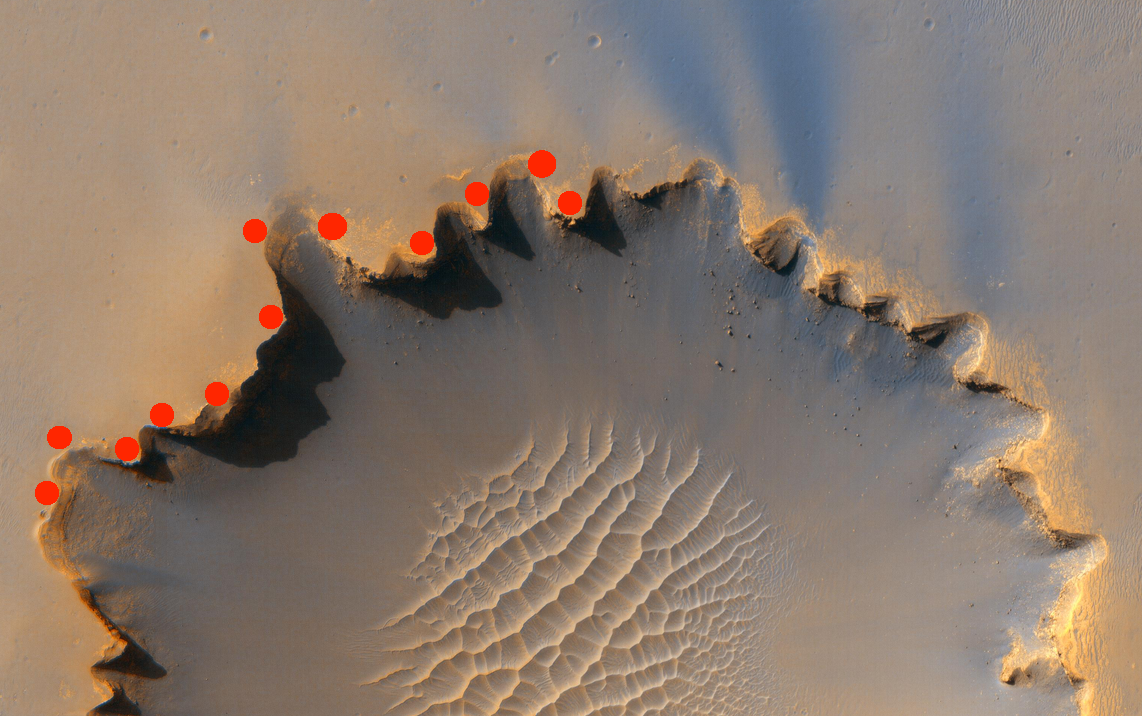} }}%
	\qquad
	\subfloat[]{{\includegraphics[width=0.74\columnwidth]{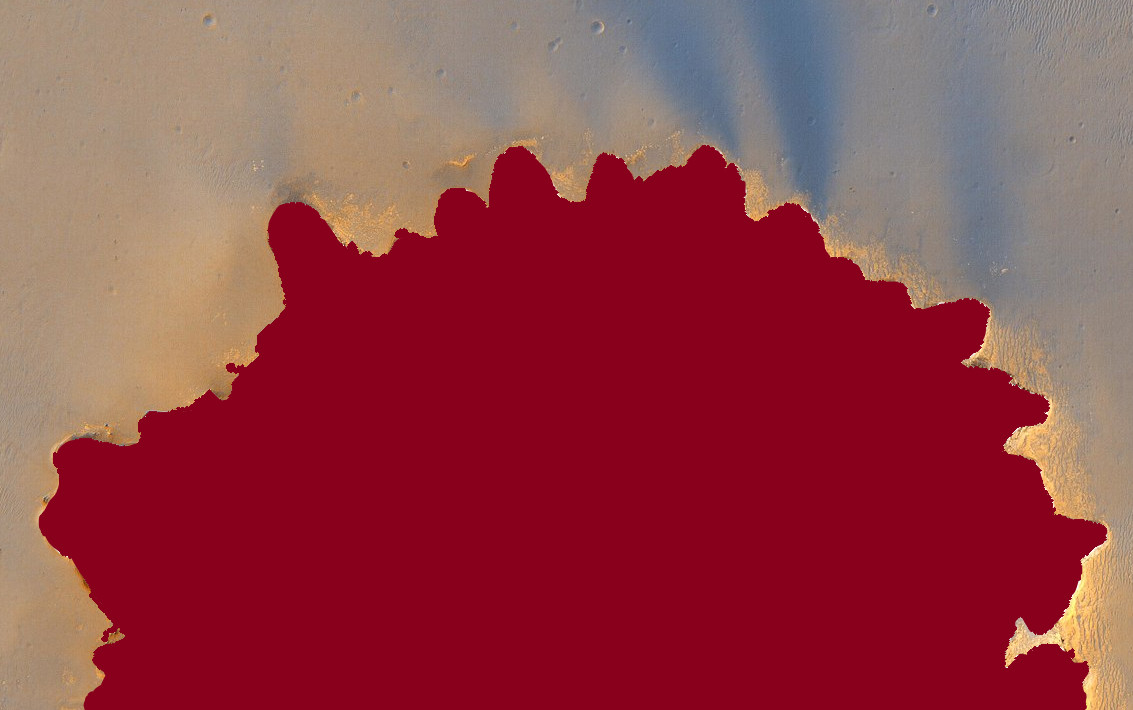} }}%
	\caption{(a) Victoria crater, (b) Opportunity rover mission traverse map, (c) replicated points of interest, and (d) unsafe area. (Image courtesy of NASA, JPL, Cornell University, and University of Arizona)}
	\label{Vic}%
\end{figure}

Victoria crater (Fig. \ref{Vic}. a) is an impact crater and is located near the equator of Mars. The crater is approximately 800 meters in diameter and it has a distinctive shape to its rim. Layered sedimentary rocks are exposed along the wall of the crater, providing invaluable information about the ancient surface condition of Mars. Since January 2004, the well-known Mars rover Opportunity had been operating around the crater and its mission path is given in Fig. \ref{Vic}. b. Opportunity worked nearly 15 years on Mars and found dramatic evidence that long ago Mars was wetter and it could have sustained microbial life, if any existed.

The scenario of interest is to train a deep neural network that can autonomously accomplish a safety-critical complex task on Mars by accessing surface images. We start with the images of the surface of Mars and given mission tasks in the form of LTL properties. We then convert the LTL properties into their corresponding LDBAs so that we can feed them into the modular deep RL algorithm. For each actor and critic network we used a feedforward net with 2 hidden layers and 400 ReLu units in each layer.

Presumably, from orbiting satellite data, we assume that the highest possible disturbance caused by different factors (such as sand storms) on the rover motion is known. This assumption can be set to be very conservative given the fact that there might be some unforeseen factors that was not captured by the satellite.

\subsection{MDP structure}
For each image, let its entire area be the MDP state space $\mathcal{S}$, where the rover location is a single state $s \in \mathcal{S}$. At each state $s \in \mathcal{S}$, the rover has a continuous range of actions $\mathcal{A}=[0,2\pi)$: 
when the rover takes an action it moves to another state (e.g., $s'$) towards the direction of the action and within a range that is randomly drawn from $(0,D]$, unless the rover hits the boundary of the image which forces the rover to remain on the boundary. 

Note that in the first experiment (Fig. \ref{Melas}), when the rover is deployed to its real mission, the precise landing location is not known. Therefore, we should encompass some randomness in the initial state $ s_0 $. However, in the second experiment (Fig. \ref{Vic}) the rover has already landed and it starts its mission from a known and fixed point. 

The dimensions of the area of interest in Fig. \ref{Melas} are $ 456.98\times 322.58 $ km and in Fig. \ref{Vic} are $ 746.98 \times 530.12 $ m. Other parameters in this numerical example have been set as $D=2$ km for Melas Chasma, $D=10$ m for the Victoria crater. We used the satellite image itself as the input MDP for the algorithm where it specifies the precise labelling assignment $L$.

\subsection{Specifications}

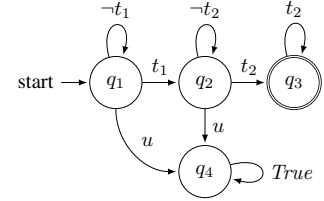
\begin{figure}[!t]\centering
\scalebox{.79}{
\begin{tikzpicture}[shorten >=1pt,node distance=1.5cm,on grid,auto] 
\node[state,initial] (q_1)   {$q_1$}; 
\node[state] (q_2) [right=of q_1] {$q_2$}; 
\node[state] (q_5) [below=of q_2] {$q_4$}; 
\node[state,accepting] (q_3) [right=of q_2] {$q_3$}; 
\path[->] 
(q_1) edge [loop above] node {$\neg t_1$} ()   	
(q_1) edge  node {$t_1$} (q_2)
(q_2) edge [loop above] node {$\neg t_2$} ()
(q_2) edge node {$t_2$} (q_3)
(q_3) edge [loop above] node {$t_2$} () 
(q_1) edge [bend right=45] node {$u$} (q_5)
(q_2) edge node {$u$} (q_5)
(q_5) edge [loop right] node {$\mathit{True}$} ();
\end{tikzpicture}
}
\caption{LDBA expressing the LTL formula in  (\ref{omega}).}
\label{Melas_a}
\end{figure}

\begin{figure*}[!t]
\centering
{{\hspace{0cm}
\scalebox{.65}{
\begin{tikzpicture}[shorten >=1pt,node distance=2cm,on grid,auto] 
\node[state,initial] (q_1)   {$q_1$}; 
\node[state] (q_2) [right=of q_1] {$q_2$};
\node[state] (q_3) [right=of q_2] {$q_3$};
\node[state] (q_4) [right=of q_3] {$q_4$};
\node[state] (q_5) [right=of q_4] {$q_5$};
\node[state] (q_6) [right=of q_5] {$q_6$};
\node[state] (q_7) [right=of q_6] {$q_7$};
\node[state] (q_8) [right=of q_7] {$q_8$};
\node[state] (q_9) [right=of q_8] {$q_9$};
\node[state] (q_10) [right=of q_9] {$q_{10}$};
\node[state] (q_11) [right=of q_10] {$q_{11}$};
\node[state] (q_13) [below= 4cm of q_6] {$q_{13}$}; 
\node[state,accepting] (q_12) [right=of q_11] {$q_{12}$}; 
\path[->] 
(q_1) edge [loop above] node {$\neg t_1$} ()   	
(q_1) edge  node {$t_1$} (q_2)
(q_2) edge [loop above] node {$\neg t_2$} ()
(q_2) edge node {$t_2$} (q_3)
(q_3) edge node {$t_3$} (q_4)
(q_4) edge node {$t_4$} (q_5)
(q_5) edge node {$t_5$} (q_6)
(q_6) edge node {$t_6$} (q_7)
(q_7) edge node {$t_7$} (q_8)
(q_8) edge node {$t_8$} (q_9)
(q_9) edge node {$t_9$} (q_10)
(q_10) edge node {$t_{10}$} (q_11)
(q_11) edge node {$t_{11}$} (q_12)
(q_3) edge [loop above] node {$\neg t_3$} () 
(q_4) edge [loop above] node {$\neg t_4$} ()
(q_5) edge [loop above] node {$\neg t_5$} ()
(q_6) edge [loop above] node {$\neg t_6$} ()
(q_7) edge [loop above] node {$\neg t_7$} ()
(q_8) edge [loop above] node {$\neg t_8$} ()
(q_9) edge [loop above] node {$\neg t_9$} ()
(q_10) edge [loop above] node {$\neg t_{10}$} ()
(q_11) edge [loop above] node {$\neg t_{11}$} ()

(q_1) edge [bend right] node {$u$} (q_13)
(q_2) edge [bend right] node {$u$} (q_13)
(q_3) edge [bend right] node {$u$} (q_13)
(q_4) edge [bend right] node {$u$} (q_13)
(q_5) edge [bend right] node {$u$} (q_13)
(q_6) edge node {$u$} (q_13)
(q_7) edge [bend left] node {$u$} (q_13)
(q_8) edge [bend left] node {$u$} (q_13)
(q_9) edge [bend left] node {$u$} (q_13)
(q_10) edge [bend left] node {$u$} (q_13)
(q_11) edge [bend left] node {$u$} (q_13)
(q_12) edge [loop above] node {$t_{12}$} ()
(q_13) edge [loop below] node {$\mathit{True}$} ();
\end{tikzpicture}} }}%
\caption{LDBA expressing the LTL formula in (\ref{ldbaeq2}).}%
\label{f3}%
\end{figure*}
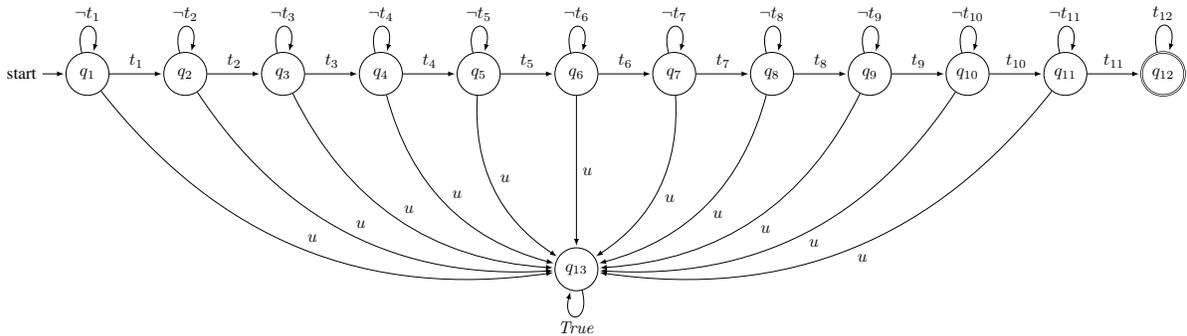

The first control objective over Melas Chasma (Fig. \ref{Melas}) is expressed by the following LTL formula:
\begin{equation}
\label{omega}
\lozenge(t_1 \wedge \lozenge t_2) \wedge \square(t_2 \rightarrow \square t_2) \wedge \square(u \rightarrow \square u),   
\end{equation} 
where $t_1$ stands for ``target 1", $t_2$ stands for ``target 2" and $u$ stands for ``unsafe" (the red region in the figure). Target 1 corresponds to the RSL (blue dots) on the right with a lower risk of the rover going to unsafe region, whereas the target 2 label goes on the left RSL that are a bit riskier to explore. Conforming to (\ref{omega}) the rover has to visit any of the right dots at least once and then proceed to the left dots, while avoiding unsafe areas. Note that according to $ \square(u \rightarrow \square u) $ in \eqref{omega} the agent can enter the unsafe area $ u $ (by climbing up the slope) but it is not able to come back due to the risk of falling. From (\ref{omega}) we build the associated B\"uchi automaton as in Fig.~\ref{Melas_a}. 

The mission task for the Victoria crater is expressed by the following LTL formula:
\begin{align}
\begin{aligned}
&\Diamond(t_1\wedge\Diamond (t_2 \wedge\Diamond(t_3 \wedge\Diamond(t_4 \wedge\Diamond (... \wedge\Diamond(t_{12})))) \wedge \\
&\square(t_{12} \rightarrow \square t_{12}) \wedge \square(u \rightarrow \square u),
\end{aligned}
\label{ldbaeq2}
\end{align}
where $t_i$ represent the ``$i$-th target", and $u$ represents ``unsafe". The $i$-th target in Fig. \ref{Vic}. c is the $i$-th red circle from the bottom left along the crater rim. According to (\ref{ldbaeq2}) the rover is required to visit the checkpoints from the bottom left to the top right sequentially, while not falling into the crater, mimicking the actual path in Fig. \ref{Vic}. b. From (\ref{ldbaeq2}), we can build the associated B\"uchi automaton as shown in Fig \ref{f3}.

\subsubsection{Experimental Outcomes}

\begin{figure}[!b]
	\centering
	\subfloat[Landing coordinates (2,2)]{{\includegraphics[width=0.4\columnwidth]{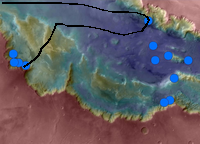} }}\hspace{2em}
	\subfloat[Landing coordinates (14,165)]{{\includegraphics[width=0.4\columnwidth]{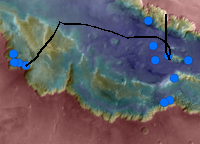} }}%
	\qquad
	\subfloat[Landing coordinates (113,199)]{{\includegraphics[width=0.4\columnwidth]{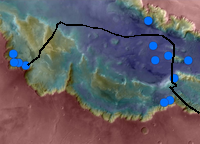} }}\hspace{2em}
	\subfloat[Landing coordinates (122,113)]{{\includegraphics[width=0.4\columnwidth]{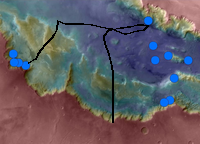} }}%
	\caption{Generated paths in the Melas Chasma experiment.}%
	\label{MDPs}%
\end{figure}

All simulations have been carried out on a machine with an Intel Xeon 3.5GHz processor and 16GB of RAM, running Ubuntu 18. In the first experiment we have employed 4 DDPG actor critic neural networks and ran simulations for 10,000 episodes. We have then tested the trained network for all safe starting position across 200 runs. Our algorithm has achieved a {success rate} of 98.8\% across 18,202 landing positions. Fig. \ref{MDPs} gives the path generated by our algorithm. Fig. \ref{MDPs}. c is particularly interesting, as we have observed a {sudden} turn before reaching the first RSL, which shows that the proposed algorithm is able to optimally learn complex policies than just smooth curve lines when needed.

\begin{figure}[!b]\centering
\includegraphics[width=0.7\columnwidth]{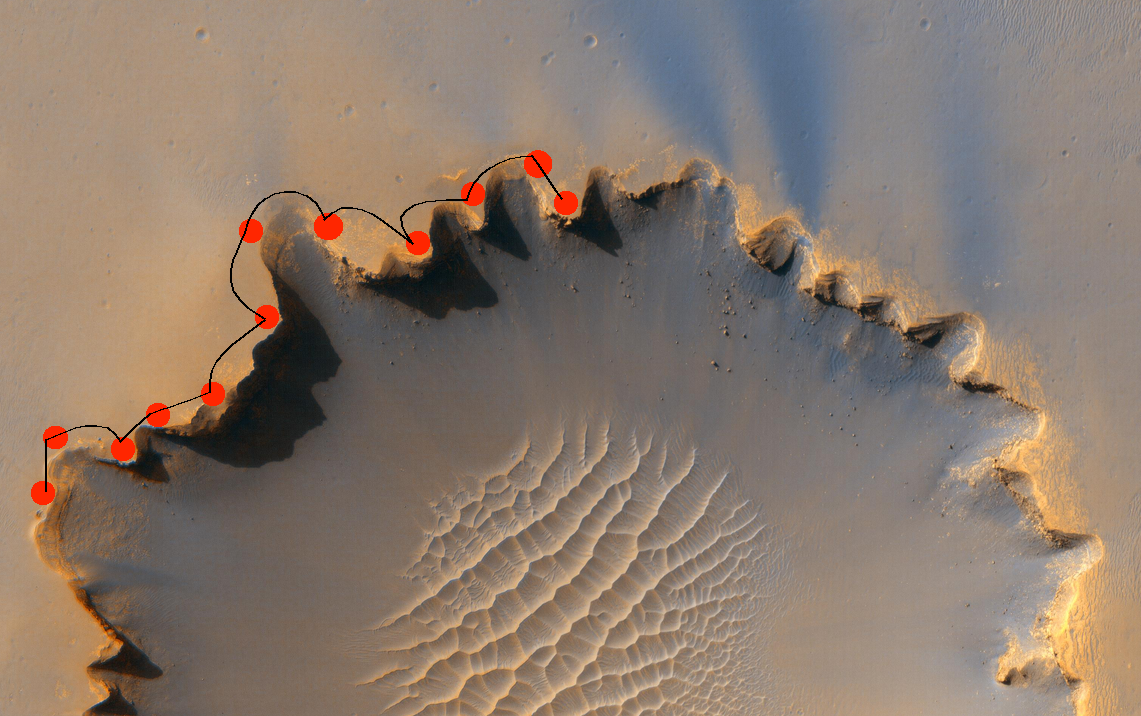}
\caption{Generated path around the Victoria crater.}
\label{Victoria_semi}%
\end{figure}

In the second experiment we have used 13 DDPG actor critic neural networks. We have ran simulations for a total of 17,000 episodes, at which point it had already converged. The training has taken approximately 5 hours to complete. We have then tested the trained network across 200 runs. Our algorithm has achieved a {success rate} of 100\% across all runs starting from $t_1$. Figure \ref{Victoria_semi} shows a generated path: we observe that the path is mostly curved away from the crater. This is due to the presence of a negative reward, as described before. We find that the negative reward is essential and that the algorithm is otherwise unable to travel from $t_7$ to $t_8$ {(around the Bottomless Bay in Fig. \ref{Vic}. b)} without introducing this negative reward. {Without the negative reward the agent insists on reaching $t_8$ via the shortest path during the exploration, resulting in constantly falling into the crater.}

In both experiments, our proposed algorithm was able to automatically modularise the given LTL task. According to Definition \ref{nonacc}, in Fig.\ref{Melas_a} $\mathds{N}=q_4$, and in Fig.\ref{f3}, $\mathds{N}=q_{13}$. The set $\mathds{A}$ guides the agent by rewarding the agent towards satisfaction of the LTL property. Hence, the general complex task can be divided into incremental sub-tasks.

\begin{table}[!t]
	\centering
	\caption{Success Rate}
	\scalebox{.8}{
	\begin{tabular}{|c|c|c|c|}
		\hline
		Case Study & Algorithm & Success Rate & Fail Rate \\
		\hline 
		\hline  
		\rule{0pt}{1em}
		\multirow{2}{*}{Melas Chasma} & Standard DDPG & 21.4\% & 78.6\% 
		\\ \cline{2-4}
		\rule{0pt}{1em}
		& Our Algorithm & 98.8\% & 1.2\%
		\\ \hline 
	\end{tabular}}
	\label{success_stats}
\end{table}

We have tried to use stand-alone DDPG as a baseline, however, the performance was too poor and the agents was unable to navigate efficiently regardless of the number of episodes. In the first experiment, the rate of trials ending in unsafe (fail) state $q_4$ and rate of trials in which the agent finds a path satisfying the LTL specification are given in Table \ref{success_stats}. Statistics are taken over 200 different starting positions after training 10,000 episodes. In the second experiment, in order to replicate the mission of the Mars rover Opportunity we fixated the starting position. As mentioned before, after training, our algorithm was able to satisfy the LTL property with success rate of 100\% across 200 trials. After the same number of training episodes, standard DDPG failed to synthesise a stable policy with LTL-satisfying traces. We are also unaware of any literature that can provide a one-shot learning baseline for such complex sequential task. Implementation details are available in the appendix.

\section{Conclusion}
In this paper we have discussed the first deep RL scheme for hierarchical continuous-state continuous-action decision making problems with temporal constraints. These problems are composed of interrelated sub-problems, that in turn might have their own sub-problems. Although the optimal decision making for each sub-problem can be effortlessly done, the original problem is quite hard to be tackled holistically, even with  state-of-the-art techniques. We have employed LTL to specify these interrelations and to assist the agent to find an optimal policy in a one-shot learning scheme. 

\vfill
\bibliographystyle{aaai}
\bibliography{sample}

\begin{thebibliography}{}

\bibitem[\protect\citeauthoryear{Abate and
  Soudjani}{2015}]{abate2015quantitative}
Abate, A., and Soudjani, S. E.~Z.
\newblock 2015.
\newblock Quantitative {A}pproximation of the {P}robability {D}istribution of a
  {M}arkov {P}rocess by {F}ormal {A}bstractions.
\newblock {\em Logical Methods in Computer Science} 11.

\bibitem[\protect\citeauthoryear{Abate \bgroup et al\mbox.\egroup
  }{2010}]{abate2010approximate}
Abate, A.; Katoen, J.-P.; Lygeros, J.; and Prandini, M.
\newblock 2010.
\newblock Approximate model checking of stochastic hybrid systems.
\newblock {\em European Journal of Control} 16(6):624--641.

\bibitem[\protect\citeauthoryear{Andreas, Klein, and Levine}{2017}]{modular}
Andreas, J.; Klein, D.; and Levine, S.
\newblock 2017.
\newblock Modular multitask reinforcement learning with policy sketches.
\newblock In {\em Proceedings of the 34th International Conference on Machine
  Learning-Volume 70},  166--175.

\bibitem[\protect\citeauthoryear{Athiwaratkun \bgroup et al\mbox.\egroup
  }{2018}]{sw}
Athiwaratkun, B.; Finzi, M.; Izmailov, P.; and Wilson, A.~G.
\newblock 2018.
\newblock Improving consistency-based semi-supervised learning with weight
  averaging.
\newblock {\em arXiv preprint arXiv:1806.05594} 2.

\bibitem[\protect\citeauthoryear{Baier and Katoen}{2008}]{bible}
Baier, C., and Katoen, J.-P.
\newblock 2008.
\newblock {\em Principles of model checking}.
\newblock MIT press.

\bibitem[\protect\citeauthoryear{Bertsekas and Shreve}{2004}]{shreve}
Bertsekas, D.~P., and Shreve, S.
\newblock 2004.
\newblock {\em Stochastic optimal control: the discrete-time case}.
\newblock Athena Scientific.

\bibitem[\protect\citeauthoryear{Bertsekas and Tsitsiklis}{1996}]{NDP}
Bertsekas, D.~P., and Tsitsiklis, J.~N.
\newblock 1996.
\newblock {\em Neuro-dynamic Programming}, volume~1.
\newblock Athena Scientific.

\bibitem[\protect\citeauthoryear{Br{\'a}zdil \bgroup et al\mbox.\egroup
  }{2014}]{brazdil}
Br{\'a}zdil, T.; Chatterjee, K.; Chmel{\'\i}k, M.; Forejt, V.;
  K{\v{r}}et{\'\i}nsk{\`y}, J.; Kwiatkowska, M.; Parker, D.; and Ujma, M.
\newblock 2014.
\newblock Verification of {M}arkov decision processes using learning
  algorithms.
\newblock In {\em ATVA},  98--114.
\newblock Springer.

\bibitem[\protect\citeauthoryear{Cauchi and Abate}{2019}]{stochy}
Cauchi, N., and Abate, A.
\newblock 2019.
\newblock Stochy-automated verification and synthesis of stochastic processes.
\newblock In {\em Proceedings of the 22nd ACM International Conference on
  Hybrid Systems: Computation and Control},  258--259.
\newblock ACM.

\bibitem[\protect\citeauthoryear{Daniel, Neumann, and
  Peters}{2012}]{options_h_1}
Daniel, C.; Neumann, G.; and Peters, J.
\newblock 2012.
\newblock Hierarchical relative entropy policy search.
\newblock In {\em Artificial Intelligence and Statistics},  273--281.

\bibitem[\protect\citeauthoryear{De~Giacomo \bgroup et al\mbox.\egroup
  }{2019}]{bolt}
De~Giacomo, G.; Iocchi, L.; Favorito, M.; and Patrizi, F.
\newblock 2019.
\newblock Foundations for restraining bolts: Reinforcement learning with
  {LTLf/LDLf} restraining specifications.
\newblock In {\em Proceedings of the International Conference on Automated
  Planning and Scheduling}, volume~29,  128--136.

\bibitem[\protect\citeauthoryear{Fu and Topcu}{2014}]{topku}
Fu, J., and Topcu, U.
\newblock 2014.
\newblock Probably approximately correct {MDP} learning and control with
  temporal logic constraints.
\newblock In {\em Robotics: Science and Systems}.

\bibitem[\protect\citeauthoryear{Gunter}{2003}]{natural2LTL3}
Gunter, E.
\newblock 2003.
\newblock From natural language to linear temporal logic: Aspects of specifying
  embedded systems in {LTL}.
\newblock In {\em Proceedings of the Monterey Workshop on Software Engineering
  for Embedded Systems: From Requirements to Implementation}.

\bibitem[\protect\citeauthoryear{Harris and Harris}{2010}]{onehot}
Harris, D., and Harris, S.
\newblock 2010.
\newblock {\em Digital design and computer architecture}.
\newblock Morgan Kaufmann.

\bibitem[\protect\citeauthoryear{Hasanbeig, Abate, and
  Kroening}{2019}]{logicalconstraint}
Hasanbeig, M.; Abate, A.; and Kroening, D.
\newblock 2019.
\newblock Logically-constrained neural fitted {Q}-iteration.
\newblock In {\em Proceedings of the 18th International Conference on
  Autonomous Agents and MultiAgent Systems},  2012--2014.

\bibitem[\protect\citeauthoryear{Kearns and Singh}{2002}]{options_1}
Kearns, M., and Singh, S.
\newblock 2002.
\newblock Near-optimal reinforcement learning in polynomial time.
\newblock {\em Machine learning} 49(2-3):209--232.

\bibitem[\protect\citeauthoryear{Kulkarni \bgroup et al\mbox.\egroup
  }{2016}]{options_2}
Kulkarni, T.~D.; Narasimhan, K.; Saeedi, A.; and Tenenbaum, J.
\newblock 2016.
\newblock Hierarchical deep reinforcement learning: Integrating temporal
  abstraction and intrinsic motivation.
\newblock In {\em Advances in neural information processing systems},
  3675--3683.

\bibitem[\protect\citeauthoryear{Lillicrap \bgroup et al\mbox.\egroup
  }{2015}]{DDPG}
Lillicrap, T.~P.; Hunt, J.~J.; Pritzel, A.; Heess, N.; Erez, T.; Tassa, Y.;
  Silver, D.; and Wierstra, D.
\newblock 2015.
\newblock Continuous control with deep reinforcement learning.
\newblock {\em arXiv:1509.02971}.

\bibitem[\protect\citeauthoryear{McEwen \bgroup et al\mbox.\egroup
  }{2014}]{water_on_mars}
McEwen, A.~S.; Dundas, C.~M.; Mattson, S.~S.; Toigo, A.~D.; Ojha, L.; Wray,
  J.~J.; Chojnacki, M.; Byrne, S.; Murchie, S.~L.; and Thomas, N.
\newblock 2014.
\newblock Recurring slope lineae in equatorial regions of {M}ars.
\newblock {\em Nature Geoscience} 7(1):53.

\bibitem[\protect\citeauthoryear{Mnih \bgroup et al\mbox.\egroup
  }{2013}]{minhd}
Mnih, V.; Kavukcuoglu, K.; Silver, D.; Graves, A.; Antonoglou, I.; Wierstra,
  D.; and Riedmiller, M.
\newblock 2013.
\newblock Playing atari with deep reinforcement learning.
\newblock {\em arXiv preprint arXiv:1312.5602}.

\bibitem[\protect\citeauthoryear{Mnih \bgroup et al\mbox.\egroup
  }{2015}]{deepql}
Mnih, V.; Kavukcuoglu, K.; Silver, D.; Rusu, A.~A.; Veness, J.; Bellemare,
  M.~G.; Graves, A.; Riedmiller, M.; Fidjeland, A.~K.; Ostrovski, G.; et~al.
\newblock 2015.
\newblock Human-level control through deep reinforcement learning.
\newblock {\em Nature} 518(7540):529--533.

\bibitem[\protect\citeauthoryear{Nikishin \bgroup et al\mbox.\egroup
  }{2018}]{swa}
Nikishin, E.; Izmailov, P.; Athiwaratkun, B.; Podoprikhin, D.; Garipov, T.;
  Shvechikov, P.; Vetrov, D.; and Wilson, A.~G.
\newblock 2018.
\newblock Improving stability in deep reinforcement learning with weight
  averaging.
\newblock In {\em Uncertainty in Artificial Intelligence Workshop on
  Uncertainty in Deep Learning}, ~5.

\bibitem[\protect\citeauthoryear{Nikora and Balcom}{2009}]{natural2LTL}
Nikora, A.~P., and Balcom, G.
\newblock 2009.
\newblock Automated identification of {LTL} patterns in natural language
  requirements.
\newblock In {\em Software Reliability Engineering, 2009. ISSRE'09. 20th
  International Symposium on},  185--194.
\newblock IEEE.

\bibitem[\protect\citeauthoryear{Pnueli}{1977}]{pnueli}
Pnueli, A.
\newblock 1977.
\newblock The temporal logic of programs.
\newblock In {\em Foundations of Computer Science},  46--57.
\newblock IEEE.

\bibitem[\protect\citeauthoryear{Precup}{2001}]{precup}
Precup, D.
\newblock 2001.
\newblock {\em Temporal abstraction in reinforcement learning.}
\newblock Ph.D. Dissertation, University of Massachusetts Amherst.

\bibitem[\protect\citeauthoryear{Sadigh \bgroup et al\mbox.\egroup
  }{2014}]{dorsa}
Sadigh, D.; Kim, E.~S.; Coogan, S.; Sastry, S.~S.; and Seshia, S.~A.
\newblock 2014.
\newblock A learning based approach to control synthesis of {Markov} decision
  processes for linear temporal logic specifications.
\newblock In {\em CDC},  1091--1096.
\newblock IEEE.

\bibitem[\protect\citeauthoryear{Sickert and
  K{\v{r}}et{\'\i}nsk{\`y}}{2016}]{sickert2}
Sickert, S., and K{\v{r}}et{\'\i}nsk{\`y}, J.
\newblock 2016.
\newblock {MoChiBA}: Probabilistic {LTL} model checking using
  limit-deterministic {B{\"u}chi} automata.
\newblock In {\em ATVA},  130--137.
\newblock Springer.

\bibitem[\protect\citeauthoryear{Sickert \bgroup et al\mbox.\egroup
  }{2016}]{sickert}
Sickert, S.; Esparza, J.; Jaax, S.; and K{\v{r}}et{\'\i}nsk{\`y}, J.
\newblock 2016.
\newblock Limit-deterministic {B{\"u}chi} automata for linear temporal logic.
\newblock In {\em CAV},  312--332.
\newblock Springer.

\bibitem[\protect\citeauthoryear{Silver \bgroup et al\mbox.\egroup
  }{2014}]{DPG}
Silver, D.; Lever, G.; Heess, N.; Thomas~Degris, D.~W.; and Riedmiller, M.
\newblock 2014.
\newblock Deterministic policy gradient algorithms.
\newblock {\em ICML}.

\bibitem[\protect\citeauthoryear{Soudjani, Gevaerts, and Abate}{2015}]{faust}
Soudjani, S. E.~Z.; Gevaerts, C.; and Abate, A.
\newblock 2015.
\newblock {FAUST$^2$: {F}ormal {A}bstractions of {U}ncountable-{ST}ate
  {ST}ochastic Processes}.
\newblock In {\em TACAS},  272--286.
\newblock Springer.

\bibitem[\protect\citeauthoryear{Sutton and Barto}{1998}]{sutton}
Sutton, R.~S., and Barto, A.~G.
\newblock 1998.
\newblock {\em Reinforcement learning: An introduction}, volume~1.
\newblock MIT press Cambridge.

\bibitem[\protect\citeauthoryear{Toro~Icarte \bgroup et al\mbox.\egroup
  }{2018}]{toro}
Toro~Icarte, R.; Klassen, T.~Q.; Valenzano, R.; and McIlraith, S.~A.
\newblock 2018.
\newblock Teaching multiple tasks to an rl agent using ltl.
\newblock In {\em Proceedings of the 17th International Conference on
  Autonomous Agents and MultiAgent Systems},  452--461.
\newblock International Foundation for Autonomous Agents and Multiagent
  Systems.

\bibitem[\protect\citeauthoryear{Vezhnevets \bgroup et al\mbox.\egroup
  }{2016}]{options_h_2}
Vezhnevets, A.; Mnih, V.; Osindero, S.; Graves, A.; Vinyals, O.; Agapiou, J.;
  et~al.
\newblock 2016.
\newblock Strategic attentive writer for learning macro-actions.
\newblock In {\em Advances in neural information processing systems},
  3486--3494.

\bibitem[\protect\citeauthoryear{Watkins and Dayan}{1992}]{watkins}
Watkins, C.~J., and Dayan, P.
\newblock 1992.
\newblock {Q}-learning.
\newblock {\em Machine learning} 8(3-4):279--292.

\bibitem[\protect\citeauthoryear{Wolff, Topcu, and Murray}{2012}]{wolf}
Wolff, E.~M.; Topcu, U.; and Murray, R.~M.
\newblock 2012.
\newblock Robust control of uncertain {M}arkov decision processes with temporal
  logic specifications.
\newblock In {\em CDC},  3372--3379.
\newblock IEEE.

\bibitem[\protect\citeauthoryear{Yan, Cheng, and Chai}{2015}]{natural2LTL2}
Yan, R.; Cheng, C.-H.; and Chai, Y.
\newblock 2015.
\newblock Formal consistency checking over specifications in natural languages.
\newblock In {\em Proceedings of the 2015 Design, Automation \& Test in Europe
  Conference \& Exhibition},  1677--1682.

\end{thebibliography}

\medskip
\appendix
\section*{Appendix}
\medskip
\section{General Considerations}
We start our training for each automaton state only once our buffer size becomes large enough (e.g. buffer size of 4096), as a small buffer size would introduce more correlation in the data. This seems to slow down training in the initial stage, but results in faster learning of the policies after the initial stage.

Also, we initially apply our algorithm as described with the state dimension being the $x$ and $y$ coordinates along with the automaton states, action dimension being the angle normalized to the range of $[-1, 1]$, where $-1$ represent $0$ degree and $1$ represents $2\pi$. However, we find that the learning algorithm tends to perform very poorly and there exists a tendency of predicting actions at the boundary of $-1$ and $1$. We believe that this incident occurs because the measure of angle is circular, where $0$ degree is essentially the same as $2\pi$ degree, and the learning algorithm is unaware of such property, and hence, it will be trapped in the local minimum. To resolve this issue, instead of predicting the angle for the action dimension, we instead predict the $sin$ and $cos$ of the angle. This approach works better as the $sin$ and $cos$ value of $0$ and $2\pi$ is the same, and hence the circular property is preserved. We implement this approach by simply predicting two dimension of range $[-1, 1]$ where their value is normalized such that the sum of their squares is equal to $1$ as $sin^2(x)+cos^2(x)=1$ for any $x$.

\medskip
\section{Instability of DDPG algorithm}

Training the DDPG algorithm is quite challenging, and in our case, we find that the DDPG algorithm is only able to traverse to the next automaton state successfully at approximately 60-70\% of the time. Therefore, the probability of reaching the $i$-th state is at most $0.7^{(i-1)}$. While this is not an issue for the Melas Chasma experiment (with only four states in the automaton), this causes great instability for the Victoria crater task. A solution to this problem is to stop the training for each set of DDPG nets once they become stable. However, since the DDPG net of each state of the automaton is not independent, once a DDPG net stops training, it is not able to get new updates from the DDPG net that it depends on, i.e. the next DDPG net in the automaton state. 

\cite{swa} showed that applying Stochastic Weight Averaging (SWA) \cite{sw} to DDPG can improve its stability. SWA is a technique that allows for solutions to be found with better generalisation in supervised and semi-supervised learning. SWA is based on averaging the weights collected during training with an SGD-like method. In supervised learning, the weights are collected at the end of each training epoch. \cite{sw} uses a constant or cyclical learning rate schedule to prevent the optimization to converge to a single solution and continue to explore the region of high performing networks.

In order to apply SWA to DDPG algorithms, \cite{swa} introduces frequency of updating the SWA weights. In our work, to initialize the weights we use the weights of the model that was trained until it is able to reach the next automaton state 8 times in a row. Then we apply SWA for the weights of both actor and critic networks.

\medskip
\section{Catastrophic forgetting}
Catastrophic forgetting is the act of overwriting previous knowledge about a task when a new task is learnt. While our agent manages to become stable after applying the SWA algorithm, it started to lose accuracy after 20,000 episodes. We believe that this is due to the algorithm forgetting how to avoid unsafe region as the experience set is filled with only successful runs. To resolve this issue, we increase the initial samples for the replay buffer to 16384 (i.e. $2^{14}$) samples before we start the training. In addition to that, we separate the experience set into successful and unsuccessful experience set for each automaton state. We then sample from both replay buffer at each epoch at a fix ratio to be tuned. We found that by separating the replay buffer and increasing the initial replay buffer sample, the algorithm is able to maintain stability after 20,000 episodes.
\end{document}